\documentclass[ 8 pt, journal]{IEEEtran}  
\IEEEoverridecommandlockouts                              

\usepackage{cite}
\usepackage{amsmath,amssymb,amsfonts}
\usepackage{algorithmic}
\usepackage{graphicx}
\usepackage{textcomp}
\def\BibTeX{{\rm B\kern-.05em{\sc i\kern-.025em b}\kern-.08em
    T\kern-.1667em\lower.7ex\hbox{E}\kern-.125emX}}
\usepackage{stfloats}
\usepackage{url}
\usepackage{verbatim}

\usepackage{mathtools}

\usepackage{times}
\usepackage{cite}
\usepackage{balance}
\usepackage{graphicx}
\usepackage{amsmath} 		
\usepackage{amssymb}  	
\usepackage{dsfont}
\usepackage{mathrsfs}
\usepackage{algorithmic}
\usepackage{algorithm}
\usepackage{color}
\usepackage{hyperref}
\usepackage{caption}
\usepackage{booktabs}

\begin{document}

\title{
 Benchmarking Vision-Based Object Tracking for USVs in Complex Maritime Environments
 }
 
\author{Muhayy Ud Din, Ahsan B. Bakht, Waseem Akram, Yihao Dong, Lakmal Seneviratne, and Irfan Hussain.
\thanks{ Khalifa University Center for Autonomous Robotic Systems (KUCARS), Khalifa University, United Arab Emirates.}%
}
\maketitle
\begin{abstract}
Vision-based target tracking is crucial for unmanned surface vehicles (USVs) to perform tasks such as inspection, monitoring, and surveillance. However, real-time tracking in complex maritime environments is challenging due to dynamic camera movement, low visibility, and scale variation. Typically, object detection methods combined with filtering techniques are commonly used for tracking, but they often lack robustness, particularly in the presence of camera motion and missed detections. Although advanced tracking methods have been proposed recently, their application in maritime scenarios is limited. To address this gap, this study proposes a vision-guided object-tracking framework for USVs, integrating state-of-the-art tracking algorithms with low-level control systems to enable precise tracking in dynamic maritime environments. We benchmarked the performance of seven distinct trackers, developed using advanced deep learning techniques such as Siamese Networks and Transformers, by evaluating them on both simulated and real-world maritime datasets. In addition, we evaluated the robustness of various control algorithms in conjunction with these tracking systems. The proposed framework was validated through simulations and real-world sea experiments, demonstrating its effectiveness in handling dynamic maritime conditions. 
The results show that SeqTrack, a Transformer-based tracker, performed best in adverse conditions, such as dust storms. Among the control algorithms evaluated, the linear quadratic regulator controller (LQR) demonstrated the most robust and smooth control, allowing for stable tracking of the USV. Videos and code can be found here: \textcolor{blue}{~\url{https://muhayyuddin.github.io/tracking/}}.

\end{abstract}

\section{Introduction}

The use of unmanned surface vessels (USVs) to perform maritime operations has increased significantly. It provides a variety of functionalities for applications like search and rescue operations~\cite{Wang2023}~\cite{Mansor2021}, maritime security~\cite{johnston2017marine}~\cite{jorge2019survey}, and environmental monitoring~\cite{Villa2016}~\cite{Han2021}. These vessels are becoming increasingly important in the marine industry due to their ability to operate autonomously, even in challenging conditions such as harsh weather conditions and in areas where human life is at risk. One of the critical features in these environments is autonomous target tracking, which ensures that USVs can effectively monitor and follow objects~\cite{jin2020vision}. This functionality is essential for tasks such as  inspection, data collection, and navigation.
To navigate through challenging sea conditions, a robust and reliable control system is required. Various USV control systems have been proposed such as; Proportional-Integral-Derivative  (PID), Sliding Mode Control (SMC), and Linear Quadratic Regulator (LQR) for robust navigation and tracking.

Target tracking approaches are typically divided into two main categories: filters-based and deep learning-based approaches~\cite{Lee2022}.
Filter-based tracking techniques, such as Kalman Filter or Particle Filter~\cite{xu2018}, are often applied to track under  target in relatively calm sea conditions. 
Although these techniques are computationally efficient and suitable for real-time applications, they can struggle under highly dynamic marine conditions.
Deep learning-based approaches have significantly improved target tracking. These techniques include Convolutional Neural Networks (CNNs) that effectively extract spatial features for object detection~\cite{kusuma2022}, while Recurrent Neural Networks (RNNs) capture temporal dependencies that enable reliable tracking of moving objects in unpredictable conditions~\cite{ondruska2016}. Another approach consists of Siamese networks that enable fast and efficient tracking by comparing features across frames~\cite{bertinetto2016}. Furthermore, Transformer-based techniques utilize attention mechanisms to enhance performance by focusing on key features while filtering out distractions such as reflections~\cite{chen2021}.

Although these methods show significant performance improvements in tracking, their use in real-time marine applications remains limited. Instead of trackers, object detectors, such as YOLO~\cite{liu2016ssd} and SSD~\cite{liu2016ssd} are mostly used in maritime environments in conjunction with filtering techniques for tracking~\cite{jmse10111783}. Trackers are mainly designed for static camera setups where the camera is fixed and objects move within the environment. In maritime conditions, however, the camera is mounted on a USV that experiences dynamic motion due to both the vessel's movement and the ocean waves. These disturbances result in constantly altering the target's position within the captured image frame. Furthermore, water splashes on the camera often blur the captured image, making detection and tracking significantly difficult.  
The application of these advanced tracking techniques in such dynamic real-time scenarios has not been extensively explored~\cite{Zhou2022}.

To address this challenge, this study proposes a vision-guided object-tracking framework to explore the performance of deep learning-based trackers in the maritime domain. The framework integrates state-of-the-art object-tracking algorithms with a low-level control algorithm to enable precise target tracking in complex maritime environments. Fig.~\ref{usv:front} shows the snapshots of the experiments in a real-world maritime environment. It is important to note that the framework is designed to be generalized, allowing any state-of-the-art tracker to be incorporated as a module. In this study, we benchmark the performance of five distinct trackers, each utilizing different core algorithms such as; CNNs, Siamese networks, or Transformers in real-time marine applications. The key contributions of this work are summarized as follows:
\begin{figure}[h]
\includegraphics[width=\columnwidth]{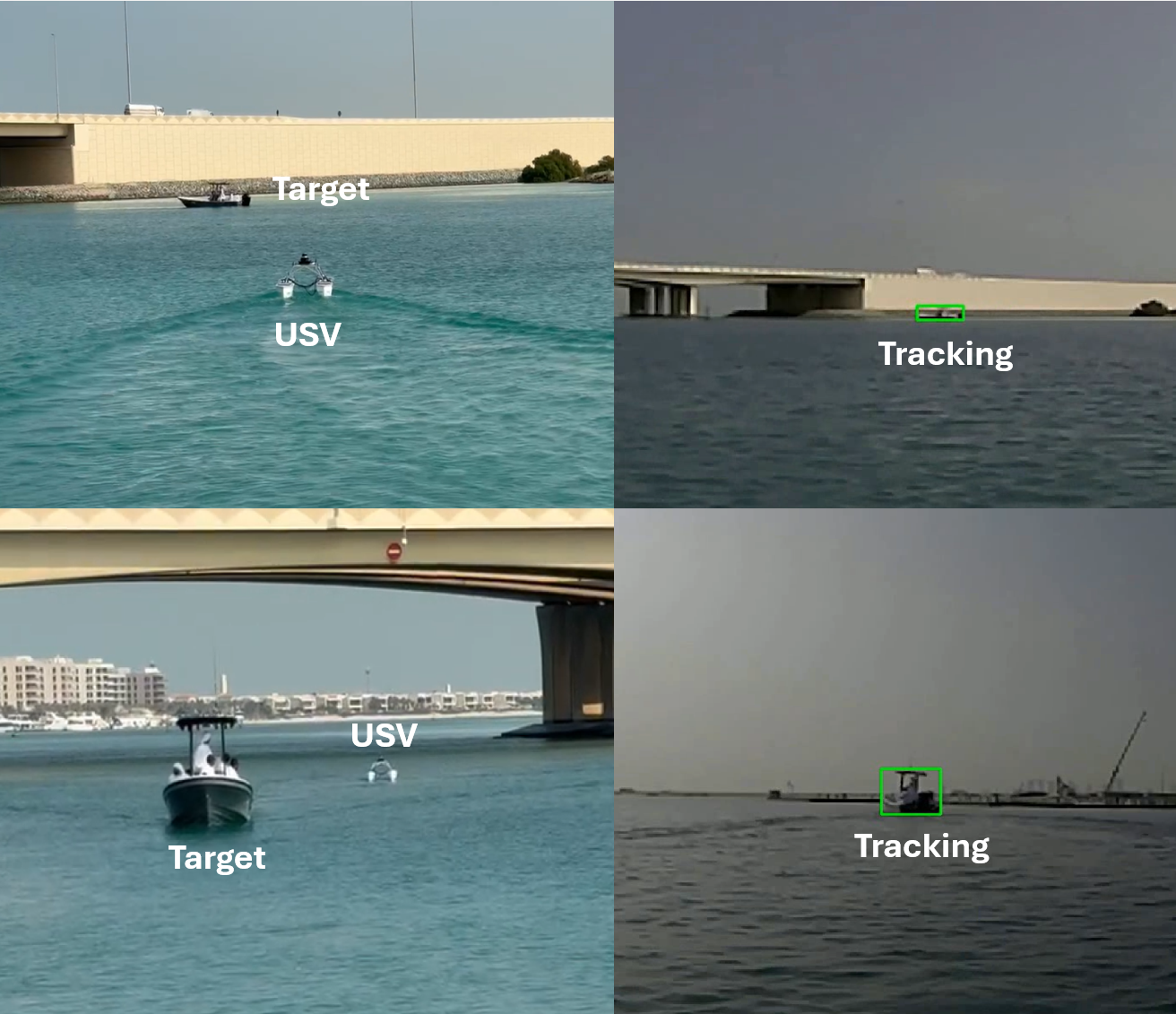}
\caption{Demonstration of Unmanned Surface Vehicle tracking a target in a real-world maritime experiment. The left images highlight the USV's navigation toward the target, while the right images show the system's tracking output.}
\label{usv:front}
\end{figure} 
\textit{Contribution:}
The key contribution of this work is a framework for vision-guided real-time object tracking specifically adapted for complex maritime environments. This framework is designed to address the specific challenges posed by dynamic and unpredictable conditions at sea. The contributions of the work are further detailed as follows:

\begin{itemize}
    \item \textit{Benchmarking of State-of-the-Art Trackers:} We perform a comprehensive evaluation of current state-of-the-art object tracking algorithms, focusing on their performance in maritime environments. This evaluation is based on a custom dataset that includes both simulated and real-world data, ensuring that the benchmarks reflect real maritime challenges such as lighting variations, reflections, and disturbances caused by waves.
    \item \textit{Integration with Control Algorithms:} The robustness of control algorithms is comprehinsively tested in combination with different tracking systems. By assessing how various trackers influence control strategies, we provide insights into optimizing navigation and object-following tasks for USV under real-world conditions.
    
    \item \textit{Validation in Simulation and Real Sea Conditions:} To ensure the robustness and applicability of the proposed framework, we validate the object tracking and control algorithms through in simulation and in real-world sea experiments. These validations demonstrate the system's ability to operate effectively in dynamic maritime environments.
\end{itemize}

\section{Related work}

Traditionally, tracking in maritime environments has relied heavily on radar-based systems~\cite{zhou2019multiple}~\cite{mitchell2017single}. While these systems are robust and reliable, however, they are very expensive and have several drawbacks, such as; their high power consumption, which is especially critical for energy-efficient USVs that operate autonomously for extended periods. In addition, radar systems tend to struggle with detecting smaller or low-reflective objects, which are common in maritime settings~\cite{zainuddin2019maritime}~\cite{Kellett2022}. Recently, vision-based tracking techniques has gained much attention as an alternative tracking approach.
The challenges faced in target tracking can be broadly categorized into four types: change in target shape, change in target scale, occlusion, disappearance of the target, and blurred image~\cite{yilmaz2006object}.
Another key challenge,
particularly in dynamic marine settings, is that the detection algorithms may fail to maintain a continuous lock on the target as it moves across frames~\cite{akram2024aquaculture}~\cite{BAKHT2024102631}, leading to missed detections and reduced reliability of target tracking using USV.

Most vision-guided object tracking systems in maritime environments for USVs utilize object detection methods combined with filtering techniques for continuous tracking~\cite{Xie2021}. Typically, object detectors such as YOLO, SSD, and RCNN are used to identify objects within a scene by computing bounding boxes, which indicate the position and size of the detected object. Once the bounding box is identified, filtering techniques, such as the Kalman filter, are applied to track the movement of the object across consecutive frames, ensuring robustness even under challenging conditions such as occlusion, noise, and camera movement. For example, ~\cite{Bewley2016} employs a faster R-CNN method for object detection. This technique combines a Kalman Filter and the Hungarian algorithm with a straightforward association metric to track bounding boxes of multiple objects in real-time. Another approach for detection and tracking in marine environment is proposed in~\cite{akram2022visual}~\cite{akram2023evaluating} which use YOLO for detection.
A fusion algorithm for object detection and tracking is proposed in~\cite{zhou2022fusion}, it used YOLO along with a kernelized correlation filter for tracking. An appraoch for tracking in extreme marine conditions is proposed in ~\cite{ahmed2023vision}. It uses YOLO with the integration of GANs for image dehazing, then PID control is used to minimize the pixel error by bringing the center of the bounding box in the center of the image. 

In recent years, vision-based tracking algorithms have gained significant attention as an alternative to traditional two-stage tracking systems. Trackers based on Siamese Networks and Transformers have demonstrated notable improvements in performance. Siamese networks, as proposed in~\cite{bertinetto2016fully}, enable fully convolutional tracking, while~\cite{li2018high} introduced high-performance tracking models. Furthermore, transformer-based models, such as ~\cite{chen2021transformer} and~\cite{cui2022mixformer}, have further advanced tracking capabilities. Despite their success, their application in real-time maritime environments, particularly for USVs, remains limited, highlighting the need for further research in this domain.

To navigate through challenging sea conditions,  various USV control systems have been proposed. For example, a Proportional Integral Derivative (PID) controller~\cite{kotian2023} was used to control the motion of the USV by establishing a dynamic model. The heading tracking controller was designed using a feedback control method to ensure accurate tracking of the desired heading based on the system's output. Vision-guided USV navigation using PID control is proposed in~\cite{din2023}. This approach uses Yolo for object detection and computes the pixel error that is translated to yaw command for the PID control.
A sliding mode control method based on the Kalman filter was proposed for USV heading control~\cite{ge2020}. The heading controller utilizes sliding mode control, incorporating a Sigmoid function to enhance the performance of the conventional asymptotic approximation rate. Linear Quadratic Regulator (LQR) has been applied to USV navigation and tracking to provide optimal control by minimizing a cost function that balances control effort and state deviation~\cite{feng2021}~\cite{yazdanpanah2013fuzzy}. LQR ensures smooth and efficient trajectory tracking while considering system constraints.
\begin{figure*}[t]
\includegraphics[width=\linewidth]{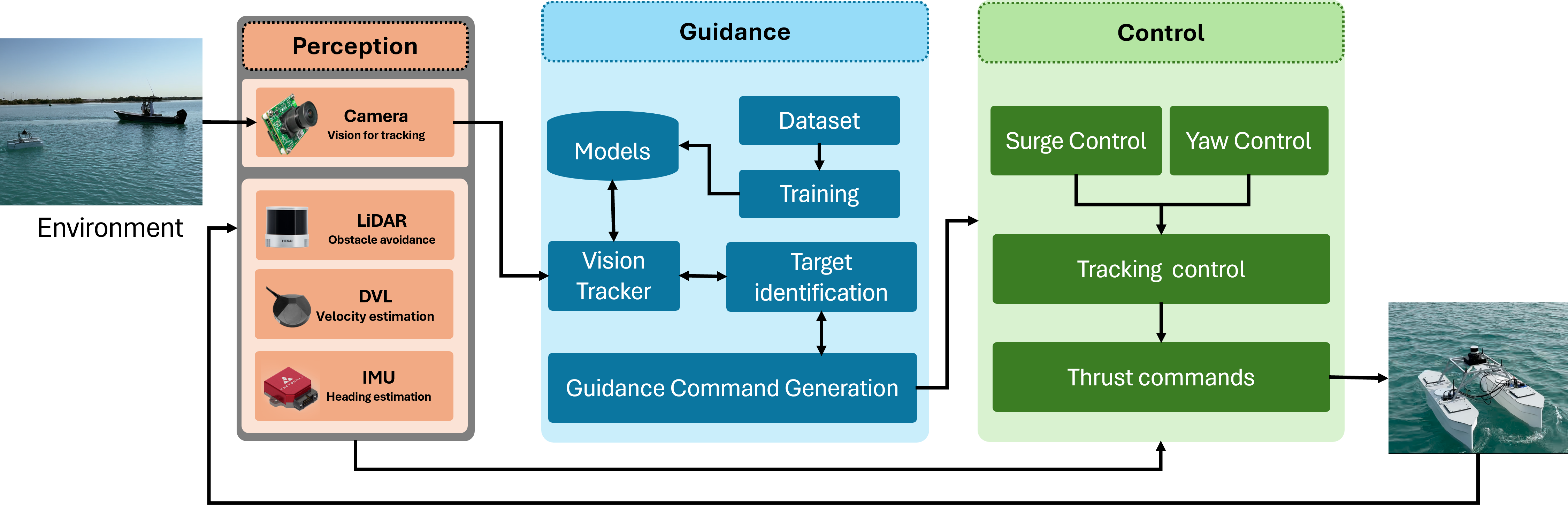}
\caption{ Proposed framework for real-time target tracking in a maritime environment. The framework is composed of three main modules: the Perception module, which incorporates sensors for environmental perception and state estimation; the Guidance module, which includes a vision-based tracker and computes guidance commands based on pixel and distance errors; and the Control module, which integrates surge and yaw control to generate tracking commands that drive the USV's thrusters, enabling precise target tracking.}
\label{fig:framework11}
\end{figure*} 

\section{Problem Formulation}
\subsection{Problem Modeling}
The problem of tracking a moving boat using a USV, equipped with a camera can be formulated in terms of pixel error in the camera frame and the physical distance between the USV and the target.
Let the position of the target boat in the camera frame be represented as:
\begin{equation}\label{tracking}
 \mathbf{p}_{\text{camera}}(t) = x_{\text{pixel}}(t), \mathcal{T}_{\text{vision}}(\mathcal{\xi}(t))  
\end{equation}\par
Where, \( \mathbf{p}_{\text{camera}}(t) = (x_{\text{pixel}}(t), y_{\text{pixel}}(t)) \) is the position of the target boat in the image, measured in pixel coordinates, \( x_{\text{pixel}}(t) \) and \( y_{\text{pixel}}(t) \) are the pixel coordinates of the target in the image at time \( t \).
 \( \xi(t) \) is the image captured by the camera at time \( t \), \( \mathcal{T}_{\text{vision}} \) is a vision-based tracking function that extracts the position of the boat from the image.
The desired position of the target in the camera frame is the center of the image, corresponding to the point \( (x_{\text{desired}}, y_{\text{desired}}) \), which is defined as:

\begin{equation}\label{p_desir}
\mathbf{p}_{\text{desired}} = (x_{\text{desired}}, y_{\text{desired}}) = \left( \frac{W}{2}, \frac{H}{2} \right) 
\end{equation}\par

Where \( W \) and \( H \) are the width and height of the image in pixels, respectively.

The pixel error between the desired and actual position of the target in the image is defined as; 

\begin{equation}\label{e_pixel}
\mathbf{e}_{\text{pixel}}(t) = \mathbf{p}_{\text{desired}} - \mathbf{p}_{\text{camera}}(t)
\end{equation}\par

The pixel error \( \mathbf{e}_{\text{pixel}}(t) \) in the camera frame is mapped to the body-fixed frame of the USV to control the yaw angle. Using the camera's intrinsic and extrinsic parameters, the transformation is represented as:
\begin{equation}\label{e_pixel}
\mathbf{e}_{\text{body}}(t) = T_{\text{camera-body}} \cdot \mathbf{e}_{\text{pixel}}(t)
\end{equation}\label{e_pixel}

Where, \( \mathbf{e}_{\text{body}}(t) = (e_x(t), e_y(t)) \) is the error in the body-fixed frame, \( T_{\text{camera-body}} \) is a transformation matrix that maps pixel error to physical coordinates in the body frame. 
The yaw angle \( \psi(t) \) is controlled based on \( e_x(t) \), which represents the lateral offset in the body frame.

The distance between the USV and the target is estimated using a LiDAR sensor. Let the distance to the target measured by the LiDAR at time \( t \) be represented as \( d_{\text{LiDAR}}(t) \). The desired distance between the USV and the target is given by; \(d_{\text{desired}} = D\), where \( D \) is the desired distance between the USV and the target.
The distance error \(e_d(t)\) is defined as:
\begin{equation}\label{e_pixel}
e_d(t) = d_{\text{desired}} - d_{\text{LiDAR}}(t)
\end{equation}\label{e_pixel}
When the target is out of the LiDAR range, which means that the target is far from the USV, it moves with a fixed forward speed \( u_{t} = u_{max}  \).  Let \( R_{\text{max}} \) be the maximum detection range of the LiDAR. 
When the target comes into the LiDAR range, i.e.,\(
d_{\text{LiDAR}}(t) \leq R_{\text{max}},\)
the USV's speed \( u(t) \) is reduced as a function of the distance to the target. The velocity is controlled to ensure that the USV maintains a safe distance of \(D\)  from the target while tracking. The velocity control function can be expressed as:

\begin{equation}\label{e_pixel}
u(t) = \begin{cases} 
      u_{\text{max}} \,  & \text{if } d_{\text{LiDAR}}(t) > R_{\text{max}}, \\
      u_{\text{max}} \cdot \frac{d_{\text{LiDAR}}(t)}{d_{\text{desired}}} & \text{if } d_{\text{LiDAR}}(t) \leq R_{\text{max}}, 
      \end{cases}
\end{equation}\label{e_pixel}

This ensures that the speed of the USV decreases as it approaches the target and maintains a fixed distance of \(D\)  from the target.

The dynamics of the USV can be modeled in general as follows:

\begin{equation}\label{e_pixel}
\mathbf{\dot{x}}(t) = f_{\text{dynamics}}(\mathbf{x}(t), \mathbf{u}(t))
\end{equation}\label{e_pixel}
Where, \( \mathbf{x}(t) = [u(t), \psi(t), r(t)] \) is the state vector consisting of surge velocity \( u(t) \), yaw angle \( \psi(t) \), and yaw rate \( r(t) \),  \( \mathbf{u}(t) = [T_L(t), T_R(t)] \) is the control input vector consisting of left and right thruster forces, \( f_{\text{dynamics}} \) is the dynamic model of the USV.

The control inputs \( \mathbf{u}(t) \) are optimized to minimize the tracking error, which is represented in terms of both pixel error and distance error. The optimization problem can be formulated as:

\begin{equation}\label{e_optimize}
\begin{aligned}
\min_{\mathbf{u}(t)} J = \int_0^T \Big( \mathbf{e}_{\text{body}}(t)^T Q_{\text{pixel}} \mathbf{e}_{\text{body}}(t) \\
+ e_d(t)^T Q_{\text{distance}} e_d(t) 
+ \mathbf{u}(t)^T R \mathbf{u}(t) \Big) dt
\end{aligned}
\end{equation}

Where,   \( Q_{\text{pixel}} \) is a weighting matrix penalizing the pixel error, \( Q_{\text{distance}} \) is a weighting matrix penalizing the distance error, \( \mathbf{u}(t) \) is the control input vector (thrust values), \( R \) is a weighting matrix penalizing the control effort.

There are certain limitations of the system that are, the thruster forces \( T_L(t) \) and \( T_R(t) \) are bounded by:
\(
T_L(t), T_R(t) \in [T_{\min}, T_{\max}]
\).
Furthermore, the surge velocity \( u(t) \) and yaw rate \( r(t) \) are subject to physical limitations.
\(
u(t) \in [u_{\min}, u_{\max}],\) \(\quad r(t) \in [r_{\min}, r_{\max}]
\).

\subsection{Problem Statement}
The objective of this study is to develope a framework for vision-guided target tracking using USV in maritime environment. The framework will be used to  identify a robust vision-based tracker \( \mathcal{T}_{\text{vision}} \) that accurately and reliably tracks a moving target (such as a boat) within the image frame captured by the USV’s onboard camera. The tracker must handle various environmental conditions, such as lighting changes, reflections, and partial occlusions, while maintaining high accuracy in detecting and tracking the target in real time.

Furthermore, the objective is to evalute the performance of various control systems to identify which controller is robust enough to handle uncertainties, such as noisy measurements from the vision system and LiDAR, and to ensure smooth navigation and tracking even in dynamic and unpredictable environments. 
At each time step \( t \), the control input \( u_t \) is applied to the USV, where \( u_t \) represents the thruster commands (surge and yaw forces) that guide the USV’s motion. The goal of control systems is to find a control input at each time step that minimizes pixel error \( e_{\text{pixel}}(t) \) and distance error \( e_d(t) \). The control optimization is formulated in terms of a cost function \( J \), shown in equ.~\ref{e_optimize}.
\subsection{Solution Overview}\label{sol:overview}

The proposed framework for real-time target tracking and benchmarking is depicted in Fig.~\ref{fig:framework11}. The framework is divided into three main modules: Perception, Guidance, and Control, each playing a critical role in effective maritime tracking.

Perception Module: This module serves as the sensory core of the system. Using various sensors, it captures and processes environmental data essential for navigation and tracking. It consists of a camera system that is responsible for the detection and continuous monitoring of targets within the maritime environment. The perception module also incorporates a LiDAR for obstacle avoidance, ensuring that the USV navigates safely by detecting and avoiding potential obstacles. A DVL and an IMU are used for USV velocity and heading estimation.

Guidance Module: is responsible for processing the data acquired by the perception system to make real-time decisions. During the offline phase, a dataset is collected for maritime tracking and various trackers are trained using this dataset (sec.~\ref{dataset} provides detail about dataset). The vision tracker actively identifies and locks onto targets, continuously adjusting its parameters based on the input received from camera (trackers detail is provided in sec.~\ref{trackers_used}). The Guidance Command Generation component takes the information processed by the vision tracker and the sensory system (LiDAR and IMU) to formulate navigation commands. These commands are computed for obstacle avoidance and to minimize errors, such as the pixel and distance error from the target, optimizing the path of the USV to follow or intercept targets effectively. 

Control Module: directly interfaces with the USV's operational hardware. It includes two key components: Surge Control and Yaw Control, which together manage the linear and angular motions of the vehicle, respectively. The Tracking Control system synthesizes the commands from the Guidance module, translating them into actionable thrust commands for the USV's thrusters. This ensures that the vehicle maintains the optimal speed and heading necessary for effective target tracking. By adjusting the thrust outputs, the USV can dynamically respond to changes in target movement or environmental conditions, as a result enhancing the effectiveness and efficiency of the maritime operation. The details of USV dynamics and control schemes is in sec~\ref{USV_dynamics} and sec~\ref{control_schemes}, respectively.

\section{Vision-based Tracking}\label{trackers_used}

Vision-based object trackers have seen significant advancements by using deep learning and optimization techniques to address challenges like occlusion, motion blur, and real-time processing. In this study, we used six state-of-the-art unique trackers such as SiamFC \cite{SiamFC}, a convolution-based tracker, ATOM \cite{ATOM} and DiMP \cite{DiMP}, which rely on correlation filters, and ToMP \cite{ToMP}, SeqTrack \cite{seqtrack}, and TaMOs \cite{TaMOs}, which employ transformer-based architectures, each chosen for its unique approach and superior performance across a range of benchmarks.
\begin{itemize}
 \item \textit{SiamFC} adopts a simple yet effective methodology for visual object tracking by framing the problem as a similarity matching task. It uses a Siamese network with two input branches: one for an exemplar image of the target and another for a search image from subsequent frames. Both inputs are processed through the same convolutional network to extract feature embeddings, ensuring that similar objects produce closely aligned representations. The core of SiamFC's methodology is the cross-correlation operation applied to these embeddings, which generates a response map indicating the target's position in the search image. This approach avoids complex model updates, relying purely on matching features in each frame to track the object over time. Due to its fully convolutional structure, SiamFC performs this process efficiently, allowing for real-time tracking while maintaining a simple workflow.

 \item \textit{ATOM} enhances visual object tracking by focusing on predicting the Intersection over Union (IoU) overlap between the target object and an estimated bounding box. It achieves this through a modulation-based network that integrates target-specific information from a reference image, leading to more precise bounding box predictions. The target estimation component is trained offline on extensive datasets, ensuring high accuracy. Additionally, ATOM features a classification module that effectively differentiates between the target and distractor objects using a two-layer fully convolutional network. To maintain real-time performance, ATOM utilizes Conjugate-Gradient-based optimization, balancing classification, estimation, and model updates.

 \item  \textit{DiMP} employs a discriminative approach to visual tracking by integrating both target and background appearance information. It uses an iterative optimization process based on a steepest descent methodology to refine the target model. DiMP also includes a module for model initialization and have the ability to learn the discriminative loss function during training, enhancing adaptability and accuracy. This end-to-end trainable framework minimizes prediction errors over time by optimizing both tracking and backbone feature extraction.

 \item \textit{ToMP} introduces a Transformer-based architecture to object tracking, replacing traditional optimization-based predictors. It employs a dual-stage approach where an initial compact target model localizes the object, and a Transformer-based predictor refines this localization by capturing global dependencies and relationships in the visual data. This method generates weights for a bounding box regressor to improve bounding box predictions based on the target state. ToMP’s end-to-end training uses both training and test frame information, allowing it to adapt and perform well across various benchmarks.

 \item \textit{SeqTrack} approaches object tracking as a sequence generation task using an encoder-decoder transformer architecture. The encoder extracts visual features from video frames, while the decoder generates bounding box coordinates autoregressively, treating them as a sequence of discrete tokens. This autoregressive process is guided by a causal mask, which ensures that each token is influenced only by previously generated tokens. By simplifying the tracking process and eliminating the need for complex classification or regression heads, SeqTrack directly predicts bounding box coordinates, offering a streamlined and efficient tracking solution.

 \item \textit{TaMOs} utilize a transformer-based approach to enhance object tracking by processing the entire frame to generate a shared feature representation for targets. It uses a fixed-size pool of object embeddings to create individual target models and incorporates a Feature Pyramidal Network (FPN) to operate on full-frame inputs. This architecture improves tracking accuracy and efficiency by leveraging global feature representations and ensuring robust performance across varying conditions. TaMOs has achieved a significant speed advantage and maintained a competitive performance on standard tracking benchmarks.
 \end{itemize}

\begin{figure}[ht]
\includegraphics[width=\columnwidth]{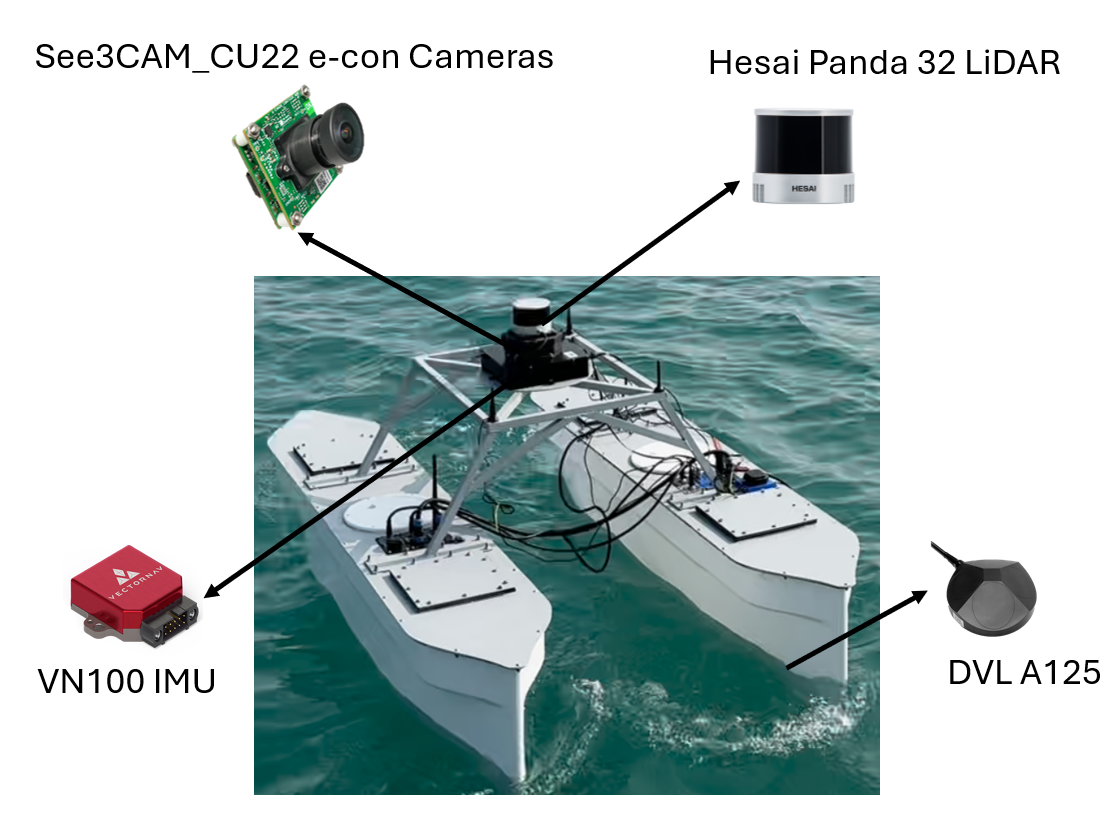}
\caption{USV used for real experiments, it is equipped with cameras, LiDAR, IMU, and DVL sensor for environment perception and state estimation.}
\label{usv}
\end{figure}

\section{USV Control}
This section provides a detailed description of the USV design, its sensor payload, and the dynamic model governing its motion. In addition, it offers an overview of the controllers employed to generate real-time control commands, allowing the USV to perform target tracking effectively.
\subsection{USV Design}
The USV features a robust catamaran design, with fiberglass hulls supported by an aluminum frame as shown in Fig.~\ref{usv}. The dimensions of the USV are 1.5 by 0.8 meters, providing a stable platform for navigation and operations. The propulsion system consists of two steerable pods, each powered by 300W, 24V thrusters, centrally installed under each hull to enable precise maneuvering and thrust control. The USV is capable of interfacing via a \texttt{CAN} bus, which enables real-time communication between various onboard systems and control units. It also supports ROS drivers, allowing seamless integration with the Robot Operating System (ROS) for efficient control and data processing. 

The sensory system of the USV includes a \texttt{Hesai Panda 32 LiDAR}, which provides high-resolution 3D mapping and obstacle detection capabilities, critical for navigation in complex environments. Furthermore, a 360-degree vision system consists of four \texttt{See3CAM\_CU22 e-con} cameras. This system enables the USV to detect and track objects effectively within its operational area. Furthermore, the \texttt{VectorNav VN100} IMU is utilized to provide reliable orientation and motion data, ensuring that the USV remains stable and responsive to control commands. The \texttt{DVL A25} measures the USV's velocity relative to the waterbed, enabling precise navigation, especially in challenging maritime environments.

To manage the computational requirements of sensor fusion, computer vision, and control systems, the USV is equipped with two advanced computing units. The \texttt{NVIDIA Xavier NX} is tasked with sensor interfacing, collecting, and processing data from the LiDAR, IMU, and cameras, while ensuring real-time communication with the ROS framework. The \texttt{Jetson Orin} handles more computationally intensive tasks, such as running complex computer vision algorithms for object detection and tracking, as well as generating control commands for navigation and maintaining a safe distance from targets. This dual-computing architecture allows the USV to perform autonomous operations effectively and efficiently, even in dynamic and unpredictable environments.

\subsection{USV Dynamic Model}\label{USV_dynamics}
The dynamics of the USV is modeled based on the forces generated by two thrusters, positioned symmetrically under the left and right hulls. The key state variables are the surge velocity \( u \), yaw angle \( \psi \), and yaw rate \( r \). The system dynamics are governed by the thrust forces \( T_L \) (left thruster) and \( T_R \) (right thruster). The kinematic equations describe how the position \( (x, y) \) and heading angle \( \psi \) evolve over time as functions of the surge velocity \( u \) and yaw rate \( r \):
\begin{equation}\label{dynamics}
    \left\{ 
        \begin{array}{l}
            \dot{x} = u \cos \psi \\
            \dot{y} = u \sin \psi \\
            \dot{\psi} = r
        \end{array}
    \right.
\end{equation}\par
Where, \( x \) and \( y \) represent the positions of the USV in the inertial frame, while \( \psi \) denotes the yaw angle, which defines the heading of the USV. The surge velocity \( u \) corresponds to the forward velocity along the body-fixed x-axis, and \( r \) is the yaw rate, representing the rate of rotation about the z-axis.

The surge motion is governed by the total thrust \( T_1 \), which is the sum of the thrust forces generated by the left and right thrusters, the total thrust and surge acceleration are computed as; 
\begin{equation}\label{dynamics}
           T_1 = T_L + T_R   \;\;\;\; \text{and} \;\;\;\;
            m \dot{u} = T_1
\end{equation}\par

This simplifies to

\begin{equation}\label{surge_acceleration}
    \dot{u} = \frac{T_L + T_R}{m}
\end{equation}
where \(\dot{u}\) represents surge acceleration.

The yaw motion (rotation around the z-axis) is governed by the differential thrust \( T_2 \), which is the difference between the thrust forces of the left and right thrusters, the differential thrust and yaw acceleration is give by; 
\begin{equation}\label{yaw_acc}
           T_2 = T_R - T_L   \;\;\;\; \text{and} \;\;\;\;
           I_{zz} \dot{r} = T_2 \cdot l
\end{equation}\par
by rearranging Equ.~\ref{yaw_acc} we can get 
  \begin{figure}[htbp]
	\centering
\includegraphics[width=0.7\columnwidth]{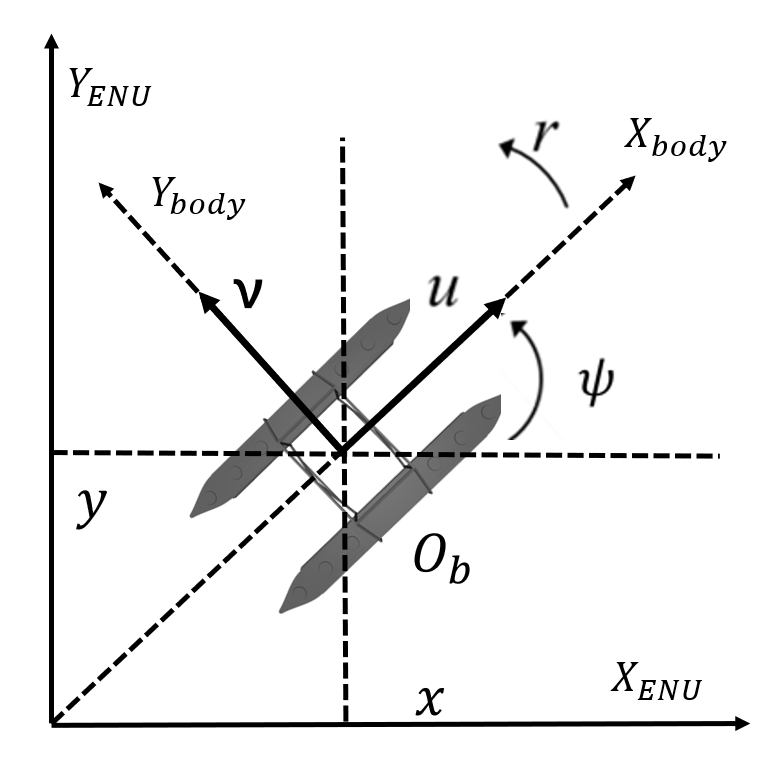}
\caption{Geometrical representation of the USV frames and associated terms, illustrating the spatial relationships and coordinate systems used for navigation and control.}\label{usv_frame} 
\end{figure}

\begin{equation}
\dot{r} = \frac{(T_R - T_L) \cdot l}{I_{zz}}
\label{rdot}
\end{equation}
Where, \( I_{zz} \) is the moment of inertia of the USV about the z-axis. \( l \) represents the distance between the center of mass of the USV and each thruster.

The combined form of Equ.~\ref{surge_acceleration} and Equ.~\ref{rdot} can be represented in state-space form. The state vector \( \mathbf{x} \) is represented as $\mathbf{x} =  [ u,  \psi, r]^T$. 
The control inputs are the total thrust \( T_1 \) and the differential thrust \( T_2 \):

\begin{equation}
\mathbf{\dot{x}} = \begin{bmatrix}
\dot{u} \\
\dot{\psi} \\
\dot{r}
\end{bmatrix}
=
\begin{bmatrix}
\frac{T_L + T_R}{m} \\
r \\
\frac{(T_R - T_L) \cdot l}{I_{zz}}
\end{bmatrix}
\end{equation}

The thrust forces generated by the left and right thrusters are related to the total thrust \( T_1 \) and differential thrust \( T_2 \) as follows:

\begin{equation}
T_L = \frac{T_1}{2} - \frac{T_2}{2}
\;\;\; \text{and} \;\;\;
T_R = \frac{T_1}{2} + \frac{T_2}{2}
\end{equation}

This ensures that the total thrust controls the forward motion (surge), while the differential thrust controls the yaw motion (heading).

The complete dynamic model of the USV, incorporating both the surge and yaw dynamics, is described by the following system of equations:
\begin{equation}\label{dynamics}
    \left\{ 
        \begin{array}{l}
            \dot{x} = u \cos \psi \\
            \dot{y} = u \sin \psi \\
            \dot{\psi} = r \\
            \dot{u} = \frac{T_L + T_R}{m} \\
            \dot{r} = \frac{(T_R - T_L) \cdot l}{I_{zz}} \\
        \end{array}
    \right.
\end{equation}\par

\begin{table}[htbp]
  \centering
  \caption{The Parameters of the USV}
  \label{table1.1}
  \begin{tabular}{|c|c|c|}
    \hline
    \textbf{Term Name} & \textbf{Terms} & \textbf{Values} \\ \hline
    Mass of the USV & $ m $ & 20  \, \text{kg}  \\ \hline
     Moment of Inertia about z-axis & $ I_{zz} $ & 3.2 \, $\text{kg} \cdot \text{m}^2$   \\ \hline
    Distance to thruster & $ l $ & 0.4 \, $\text{m}$ \\ \hline
    Surge Velocity & $ u $ & 1.5 \, $\text{m/s}$ \\ \hline
    Surge Acceleration & $ \dot{u} $ & 10 \, $\text{m/s}^2$ \\ \hline
    Yaw Acceleration & $ \dot{r} $ & 50 \, $\text{deg/s}^2$ \\ \hline
    Thrust Coefficient & $ k $ & 0.1 \\ \hline
  \end{tabular}
\end{table}

\subsection{Control Scheme for Tracking}\label{control_schemes}
This subsection will outline the control laws employed in the control schemes used for target tracking, including Proportional-Integral-Derivative (PID) control, Sliding Mode Control (SMC), and Linear Quadratic Regulator (LQR). Each of these control strategies will be described in terms of their formulation for real-time tracking.

\subsubsection{PID Control}

PID control is a widely used control strategy that adjusts the control input based on the error between the desired and actual system states. In the context of USV control, the PID controller is applied to regulate both the surge velocity \( u \) and the yaw angle \( \psi \) (heading) by adjusting the thrust forces of the left and right thrusters.

The objective of the PID controller in the surge direction is to regulate the forward velocity \( u \), such that it tracks the desired surge velocity \( u_{\text{desired}} \). The error in the surge velocity is defined as:

\begin{equation}
e_u(t) = u_{\text{desired}}(t) - u(t)
\end{equation}

The PID control law for surge is expressed as:

\begin{equation}
T_1(t) = K_{p,u} e_u(t) + K_{i,u} \int_0^t e_u(\tau) d\tau + K_{d,u} \frac{d e_u(t)}{dt}
\end{equation}

Where, \( K_{p,u} \), \( K_{i,u} \), and \( K_{d,u} \) are the proportional, integral, and derivative gains for surge control. \( e_u(t) \) is the error in surge velocity at time \( t \).

The PID controller for yaw regulates the yaw angle \( \psi \) such that the USV tracks the desired heading \( \psi_{\text{desired}} \). The yaw error is defined as:

\begin{equation}
e_\psi(t) = \psi_{\text{desired}}(t) - \psi(t)
\end{equation}

The PID control law for yaw is:

\begin{equation}
T_2(t) = K_{p,\psi} e_\psi(t) + K_{i,\psi} \int_0^t e_\psi(\tau) d\tau + K_{d,\psi} \frac{d e_\psi(t)}{dt}
\end{equation}

Where, \( K_{p,\psi} \), \( K_{i,\psi} \), and \( K_{d,\psi} \) are the proportional, integral, and derivative gains for yaw control, \( e_\psi(t) \) is the error in yaw at time \( t \).

\subsubsection{SMC Control}

SMC is a robust nonlinear control technique that forces the system states to follow a predefined sliding surface, ensuring robustness against disturbances and uncertainties.

The sliding surface for surge is defined as:

\begin{equation}
s_u = \lambda_u (u_{\text{desired}} - u) + \dot{u}_{\text{desired}} - \dot{u}
\end{equation}

The control law for surge is:

\begin{equation}
T_1 = m \left( \dot{u}_{\text{desired}} + \lambda_u (u_{\text{desired}} - u) \right) - \eta_u \, \text{sgn}(s_u)
\end{equation}

Where, \( \eta_u \) is the control gain. \( \lambda_u \) is the sliding surface coefficient.
\(\text{sgn}(s_u)\) refers to the sign function.

The sliding surface for yaw is defined as:
\begin{equation}
s_\psi = \lambda_\psi (\psi_{\text{desired}} - \psi) + \dot{\psi}_{\text{desired}} - \dot{\psi}
\end{equation}

The control law for yaw is:

\begin{equation}
T_2 = I_{zz} \left( \dot{r}_{\text{desired}} + \lambda_\psi (\psi_{\text{desired}} - \psi) \right) - \eta_\psi \, \text{sgn}(s_\psi)
\end{equation}

Where, \( \eta_\psi \) is the control gain, \( \lambda_\psi \) is the sliding surface coefficient.

\subsubsection{LQR Control}

LQR is an optimal control technique that provides a state-feedback control law for minimizing a quadratic cost function. The LQR controller aims to regulate the surge velocity \( u \) and the yaw angle \( \psi \) of the USV, minimizing the deviation from desired values while reducing the control effort. 

 The state-space representation of the system dynamics is given by:

\begin{equation}
\mathbf{\dot{x}} = A \mathbf{x} + B \mathbf{u}
\end{equation}

Where:
\[
A = \begin{bmatrix}
0 & 0 & 0 \\
0 & 0 & 1 \\
0 & 0 & 0
\end{bmatrix}, \quad B = \begin{bmatrix}
\frac{1}{m} & 0 \\
0 & 0 \\
0 & \frac{l}{I_{zz}}
\end{bmatrix}
\]

Here,  \( \mathbf{u} = \begin{bmatrix} T_1 \\ T_2 \end{bmatrix} \) represents the control inputs.

The LQR controller minimizes the following quadratic cost function:

\begin{equation}
J = \int_0^\infty \left( \mathbf{x}^T Q \mathbf{x} + \mathbf{u}^T R \mathbf{u} \right) dt
\end{equation}

Where, \( Q \) is a positive semi-definite matrix that penalizes deviations from the desired states (surge velocity, yaw angle, and yaw rate). \( R \) is a positive definite matrix that penalizes the control effort (thrust forces \( T_1 \) and \( T_2 \)). \( \mathbf{x} \) is the state vector, and \( \mathbf{u} \) is the control input vector.

The optimal control law is a state-feedback law given by:

\begin{equation}
\mathbf{u} = -K \mathbf{x}
\end{equation}

Where \( K \) is the optimal feedback gain matrix computed by solving the algebraic Riccati equation:

\begin{equation}
A^T P + P A - P B R^{-1} B^T P + Q = 0
\end{equation}

Here, \( P \) is the solution to the Riccati equation, and the feedback gain matrix \( K \) is calculated as:

\begin{equation}
K = R^{-1} B^T P
\end{equation}

Thus, the control inputs \( T_1 \) and \( T_2 \) are computed as:

\begin{equation}
\mathbf{u} = \begin{bmatrix} T_1 \\ T_2 \end{bmatrix} = -K \mathbf{x}
\end{equation}

For surge control, the LQR controller regulates the surge velocity \( u \) to track the desired velocity \( u_{\text{desired}} \). The control input for surge is the total thrust \( T_1 \), which is given by:

\begin{equation}
T_1 = -K_u \mathbf{x}
\end{equation}

Where \( K_u \) is the part of the feedback gain matrix \( K \) that corresponds to the surge dynamics. This ensures that the surge velocity \( u \) is regulated optimally, minimizing the control effort required from the thrusters.

For yaw control, the LQR controller regulates the yaw angle \( \psi \) and yaw rate \( r \). The control input for yaw is the differential thrust \( T_2 \), which is given by:

\begin{equation}
T_2 = -K_\psi \mathbf{x}
\end{equation}

Where \( K_\psi \) is the part of the feedback gain matrix \( K \) that corresponds to the yaw dynamics. This ensures that both yaw angle and yaw rate errors are minimized with optimal control effort.

\begin{figure*}[ht]
    \centering
    \includegraphics[width=\linewidth, page=1]{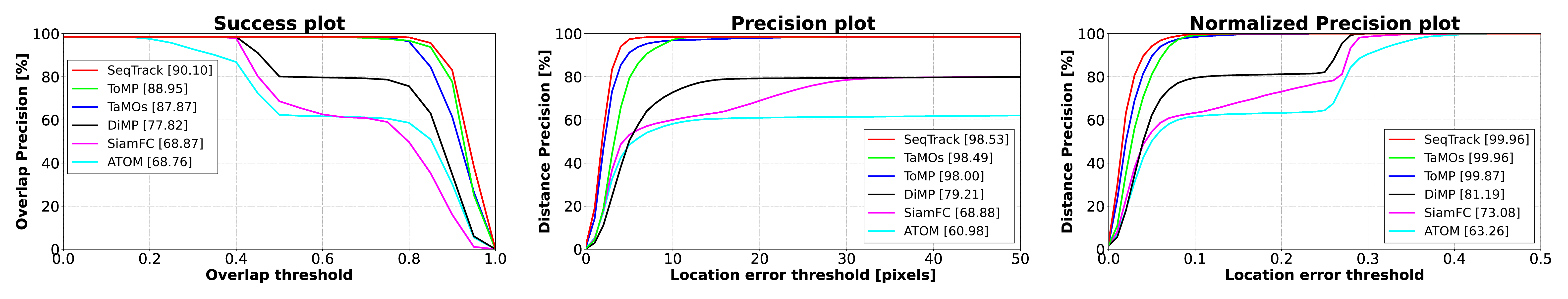} 
       \caption{Overall evaluation results on the simulation boat dataset using six trackers in terms of success, precision, and normalized precision plots.}
    \label{fig:simperformance}
\end{figure*}

\begin{figure*}[t]
    \centering
    \includegraphics[width=\linewidth, page=1]{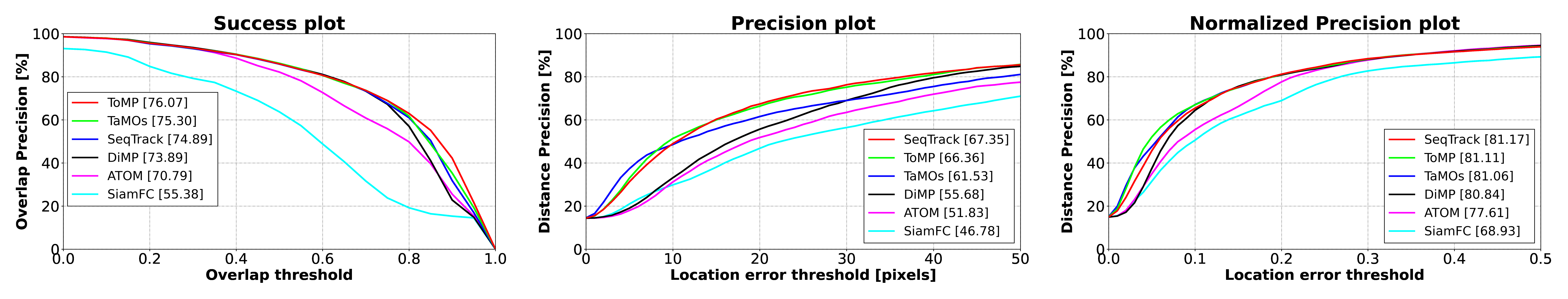} 
    \caption{Overall evaluation results on the real boat dataset using six trackers in terms of success, precision, and normalized precision plots.}
 
    \label{fig:realperformance}
\end{figure*}

\begin{table}[htbp]
    \centering
    \caption{Performance Comparison on Simulation Boat Dataset}
    \begin{tabular}{|l|c|c|c|c|c|}
        \hline
        \textbf{Tracker} & \textbf{AUC} & \textbf{OP50} & \textbf{OP75} & \textbf{Precision} & \textbf{Norm Precision} \\
        \hline
        SiamFC     & 68.87 & 68.64 & 59.01 & 68.88 & 73.08 \\
        \hline
        ATOM       & 68.76 & 62.35 & 60.60 & 60.98 & 63.26 \\
        \hline
        DiMP       & 77.82 & 80.12 & 78.66 & 79.21 & 81.19 \\
        \hline
        ToMP       & 88.95 & 98.49 & 97.45 & 98.00 & 99.87 \\
        \hline
        TaMOs      & 87.87 & 98.51 & 98.26 & 98.49 & 99.96 \\
        \hline
        SeqTrack   & \textbf{90.10} & \textbf{98.53} & \textbf{98.52} & \textbf{98.53} & \textbf{99.96} \\
        \hline
    \end{tabular}
\end{table}
\begin{table}[htbp]
    \centering
    \caption{Performance Comparison on Real Boat Dataset}
    \begin{tabular}{|l|c|c|c|c|c|c|}
    \hline
        \toprule
        \textbf{Tracker} & \textbf{AUC} & \textbf{OP50} & \textbf{OP75} & \textbf{Precision} & \textbf{Norm Precision} \\
        \hline
        \midrule
        SiamFC     & 55.38 & 63.72 & 23.78 & 46.78 & 68.93 \\
        \hline
        ATOM       & 70.79 & 82.18 & 55.88 & 51.83 & 77.61 \\
        \hline
        DiMP       & 73.89 & 86.10 & 67.26 & 55.68 & 80.84 \\
        \hline
        ToMP       & \textbf{76.07} & 85.97 & \textbf{68.96} & 66.36 & 81.11 \\
        \hline
        TaMOs      & 75.30 & \textbf{86.22} & 68.93 & 61.53 & 81.06 \\
        \hline
        SeqTrack   & 74.89 & 85.88 & 67.62 & \textbf{67.35} & \textbf{81.17} \\
        \hline
        \bottomrule
    \end{tabular}
\end{table}
\section{Trackers Benchmarking}\label{sec:benchmarking}
\subsection{Dataset}\label{dataset}
In this study we used two distinct datasets to ensure a detailed assessment of tracker's performance. The primary dataset, a real-world dataset, was captured with a camera mounted on an Unmanned Surface Vehicle (USV). This dataset consists of 21 videos recorded at 30 fps, which were carefully divided into 12 videos for training, offering a rich variety of tracking scenarios across different environmental conditions and target behaviors. The remaining 9 videos were used for testing, providing an extensive evaluation of the trackers performance in realistic and dynamic conditions. 
In addition to the real-world dataset, we also created a simulation dataset comprising of 5 videos generated within a controlled virtual environment. These simulation videos were designed to provide a separate, controlled setting for evaluating the trackers. The simulated dataset conditions are also utilized to test multiple control algorithms. 

The dataset was collected primarily under conditions common to the Saadiyat Island UAE, featuring calm and clear seas with moderate wave and wind variations. To further enrich the data set, we used synthetic data from the MBZIRC simulator to incorporate dust storms and variable winds. Although the dataset provides a controlled framework for testing and benchmarking the performance of controllers and trackers, it has some limitations, as it may not fully represent some conditions that USVs could encounter in broader maritime environments. The dataset is mainly captured during daylight; more severe conditions, such as low light conditions (after sunset), high-intensity storms, irregular sea states with complex currents, and varying temperatures, were not fully tested in real-world experiments.  This limitation highlights the area for future work; we are expanding the dataset to encompass a wider range of environmental conditions that will be tested.   Although the current data set has proven effective for tracking stability in moderate conditions. Variability and unexpected behaviors in harsh environments could affect the consistency of the observed results, especially for the PID and SMC controllers, which demonstrated higher oscillation and instability under simulated disturbances.

\subsection{Evaluation Metrics}
We utilized five metrics to evaluate the performance of trackers named as, Area Under Curve (AUC), Overlap Precision at 50\% (OP50), Overlap Precision at 75\% (OP75), Precision, and Normalized Precision. AUC measures the overall performance by evaluating the area under the precision-recall curve, providing a comprehensive summary of the algorithm's effectiveness across varying levels of precision and recall. OP50 and OP75 measure the percentage of frames where the tracker’s bounding box has at least 50\% and 75\% overlap with the ground truth, respectively. These metrics reflect the algorithm's capability to maintain accuracy across varying overlap thresholds. Precision measures the percentage of frames where the tracker successfully identifies and follows the target, thereby indicating the algorithm's accuracy in tracking. Normalized Precision adjusts the precision score relative to the highest precision achieved by any tracker, offering a comparative measure of performance. Together, these metrics provide a detailed evaluation of the tracking algorithms, highlighting their strengths and weaknesses and facilitating the selection of the most effective tracker for specific applications.

\subsection{Benchmarking on Simulation Dataset}
As shown in Table 2., we evaluated the simulation dataset by using five metrics: AUC, OP50, OP75, precision, and normalized precision. SiamFC consistently underperformed, with lower AUC and precision scores. ATOM showed notable improvement, achieving a higher AUC of 70.79 and better overlap precision, indicating robust tracking. DiMP also performed well, with an AUC of 73.89 and high overlap precision, reflecting its effectiveness. ToMP excelled with the highest precision of 66.36\% and an AUC of 76.07, showcasing superior performance in precise tracking. SeqTrack provided a balanced performance, with a solid AUC of 74.89 and the highest precision, demonstrating versatility in handling various tracking challenges. Overall as shown in Fig.~\ref{fig:simperformance}, ToMP and SeqTrack emerged as the most effective trackers, with DiMP also showing strong performance. These results highlight the strengths of each algorithm and the importance of choosing the right model based for controlling the USV.

\begin{figure}[htbp]
    \centering
    \includegraphics[width=\columnwidth]{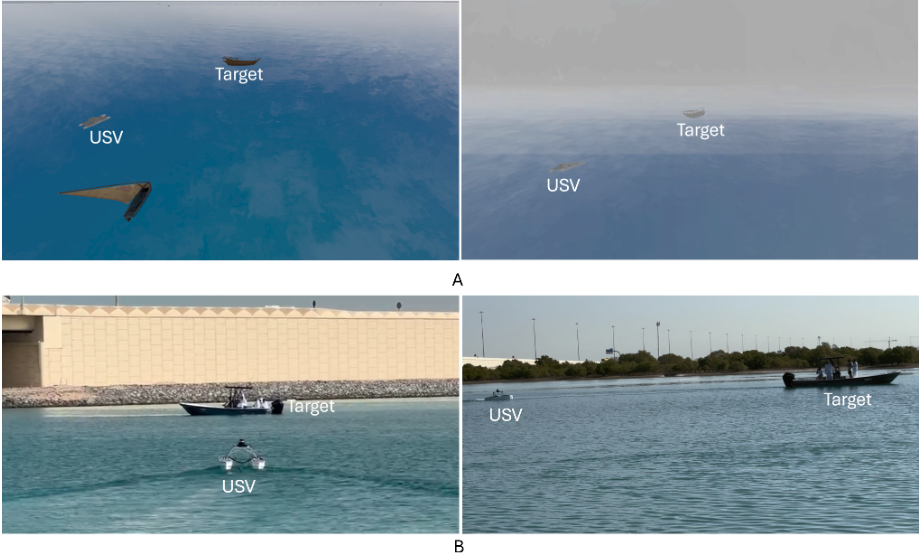} 
    \caption{Examples of simulated and real experimental setups. (A) Depicts scenarios in the simulation environment, including calm sea conditions and dust storm simulations. (B) Shows tracking scenarios from real-world experiments, highlighting the performance of the system in live maritime demonstrations.}
    \label{fig:scene}
\end{figure}

\begin{figure}[htbp]
    \centering
    \includegraphics[width=\columnwidth]{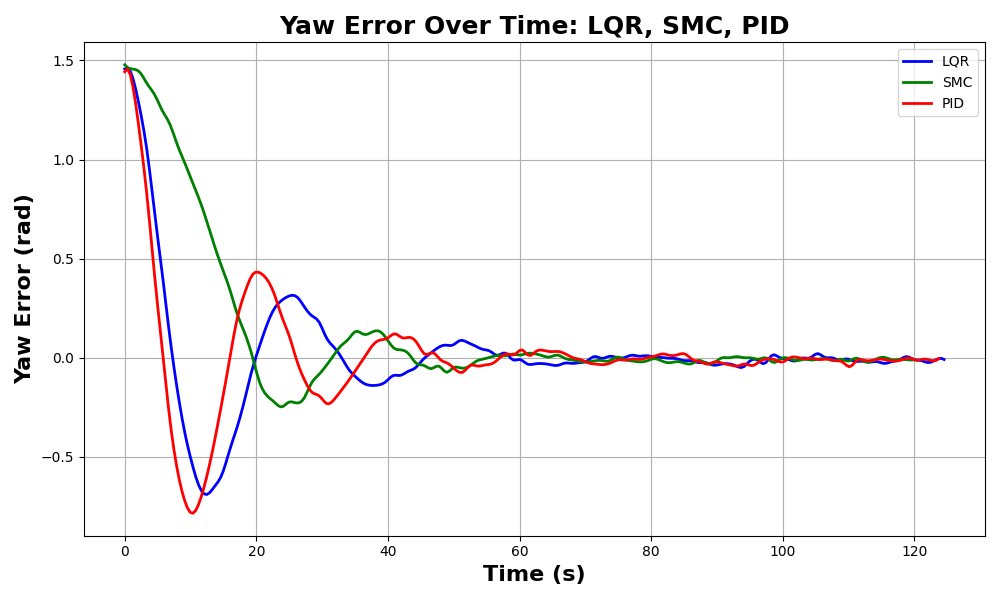} 
    \caption{Yaw error over time for USV control using PID, SMC, and LQR controllers. The PID controller (red) shows faster initial response but higher overshoot and oscillations. The SMC (green) exhibits slower convergence with smoother behavior, while the LQR (blue) provides balanced performance with moderate initial response and fewer oscillations.}
    \label{fig:tuning}
\end{figure}

\begin{figure}[t]
    \centering
\includegraphics[width=\columnwidth]{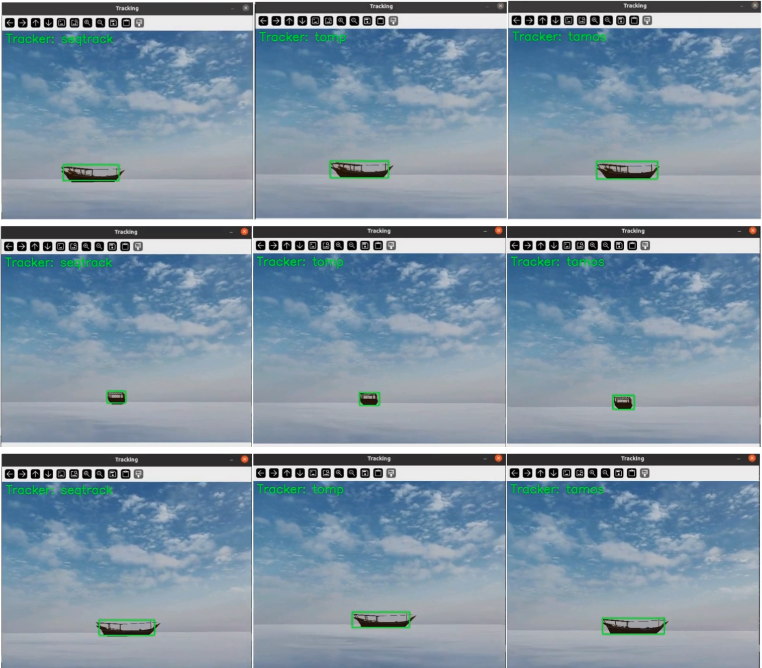} 
    \caption{Sequence of tracking snapshots using SeqTrack, ToMP, and TaMOs at varying distances and viewing angles in a clear sea environment.}
    \label{fig:clearenvironment}
\end{figure}
\begin{figure}[t]
    \centering
\includegraphics[width=\columnwidth]{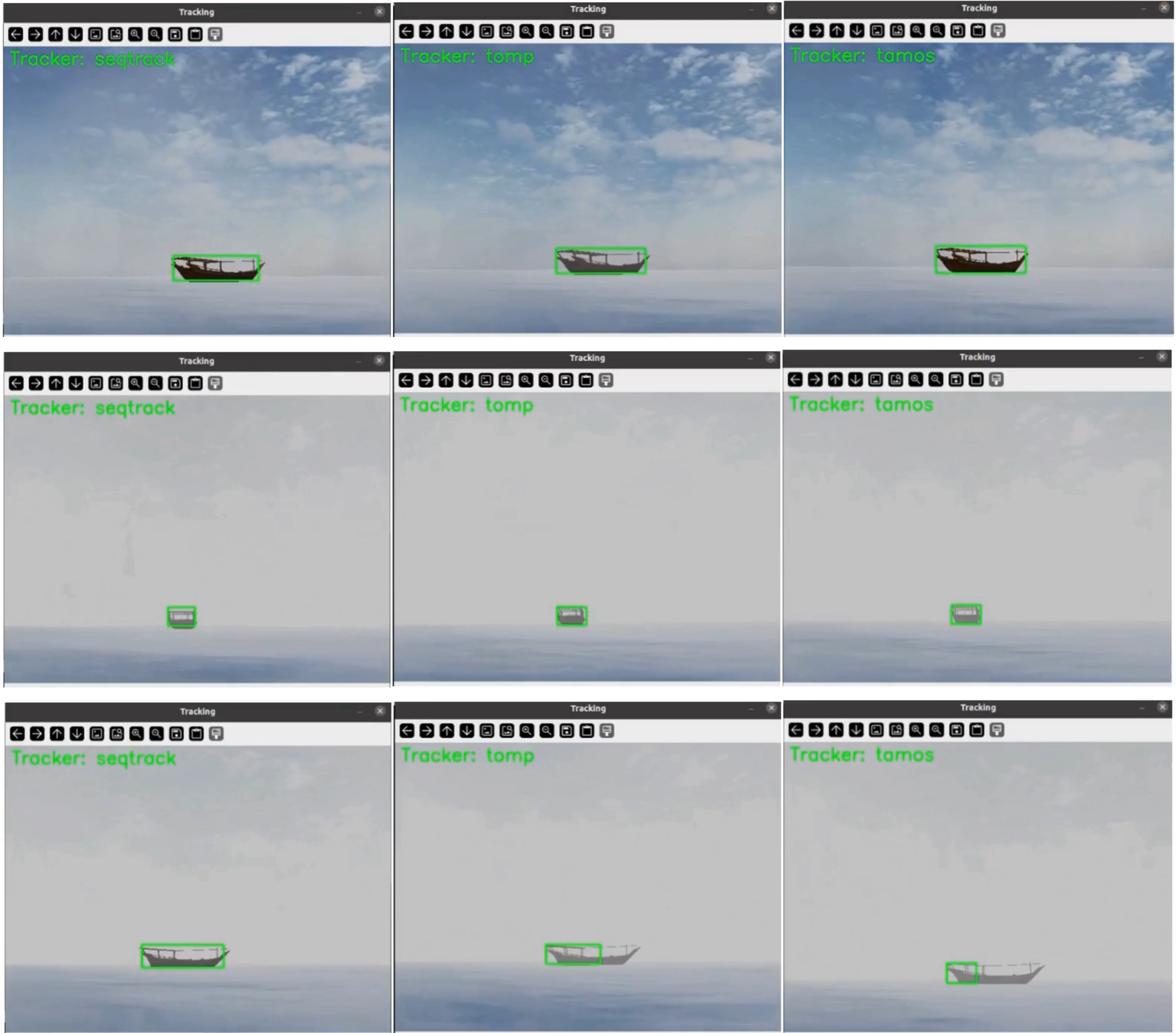} 
    \caption{Sequence of tracking snapshots using SeqTrack, ToMP, and TaMOs at varying distances and viewing angles in dust storm.}
    \label{fig:dust}
\end{figure}

\begin{figure}[t]
    \centering
    \includegraphics[width=\columnwidth, page=1]{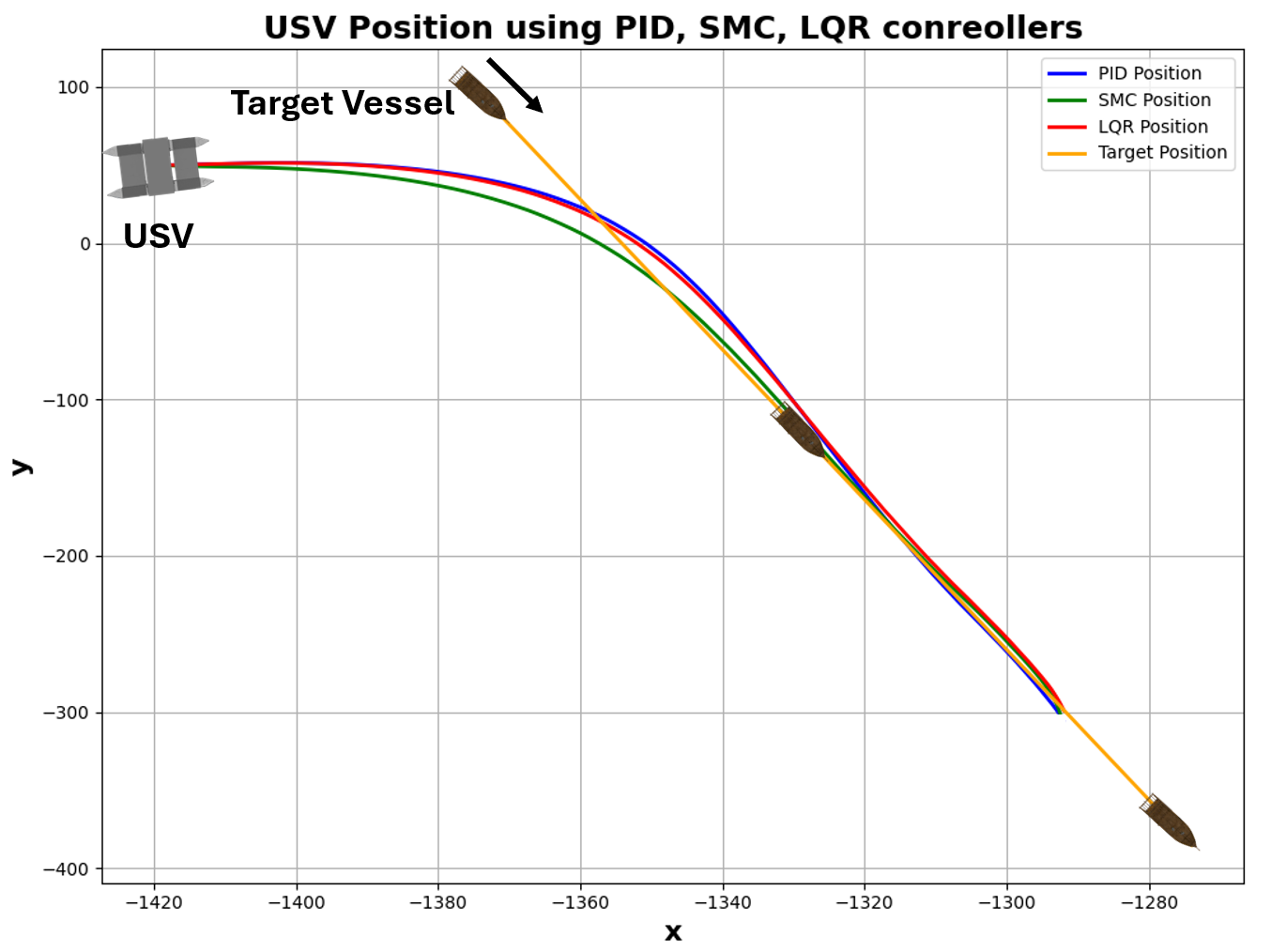} 
    \caption{Tracking a target moving in a straight line under calm and clear sea conditions using LQR, SMC, and PID controllers.}
    \label{fig:linecontrol}
\end{figure}

\begin{figure}[t]
    \centering
    \includegraphics[width=\columnwidth, page=1]{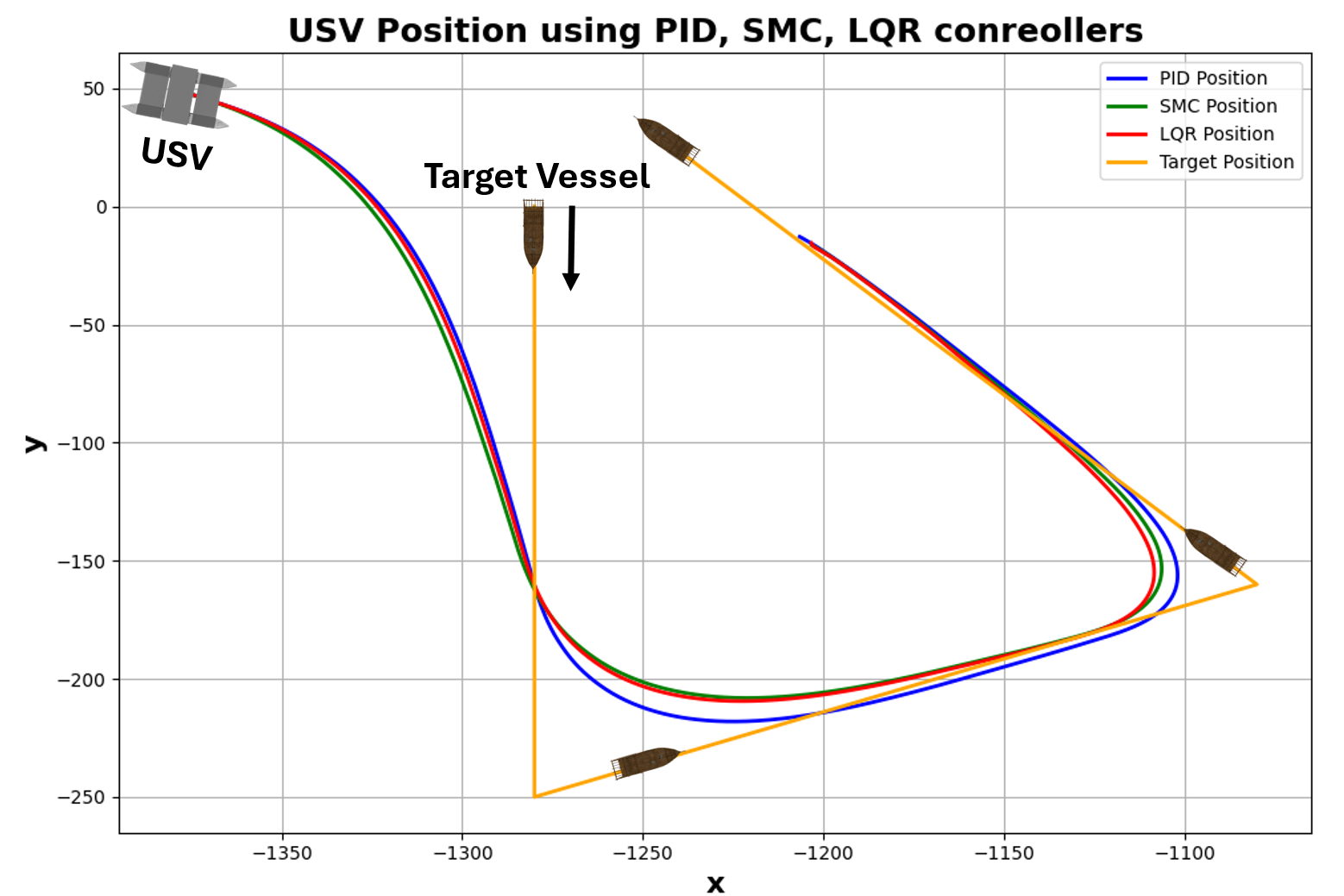} 
    \caption{Tracking a target moving on a triangular path under calm and clear sea conditions using LQR, SMC, and PID controllers.}
    \label{fig:tricontrol}
\end{figure}

\begin{figure}[t]
    \centering
    \includegraphics[width=\columnwidth, page=1]{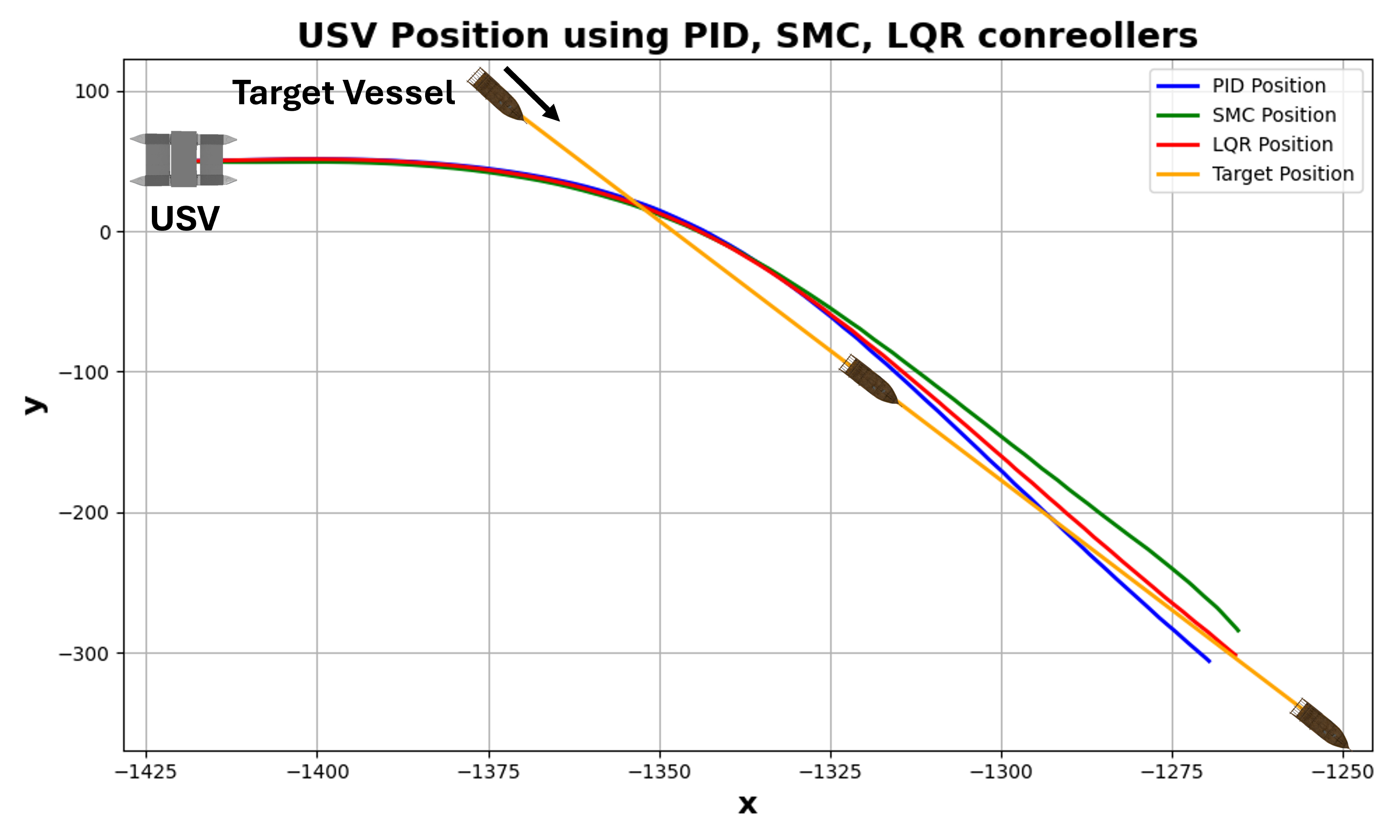} 
    \caption{Tracking a target moving in a straight line under waves and winds using LQR, SMC, and PID controllers.}
    \label{fig:dustline}
\end{figure}
\begin{figure}[t]
    \centering
    \includegraphics[width=\columnwidth, page=1]{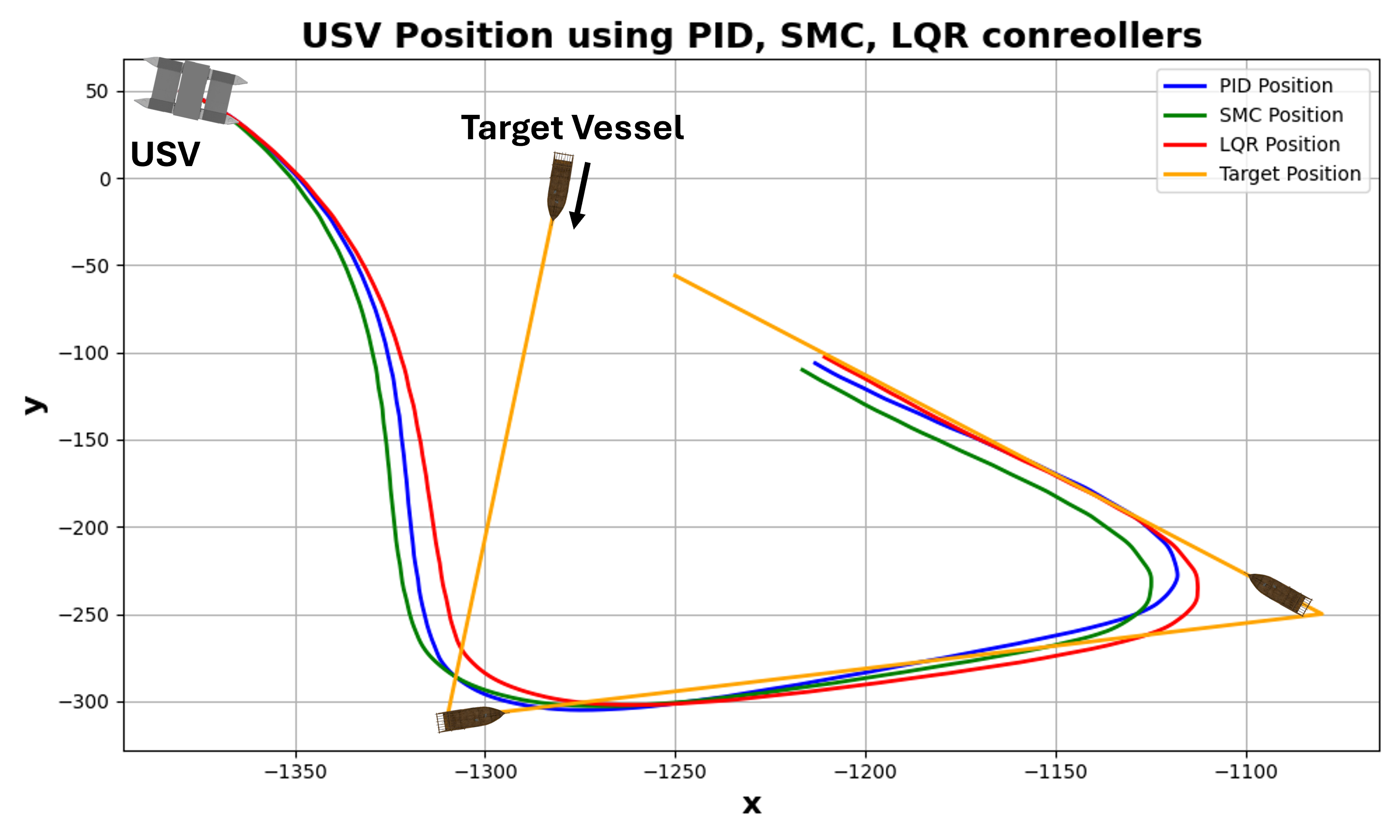} 
    \caption{Tracking a target moving on a triangular path under waves and winds using LQR, SMC, and PID controllers.}
    \label{fig:triwind}
\end{figure}

\begin{figure}[t]
    \centering
    \includegraphics[width=\columnwidth, page=1]{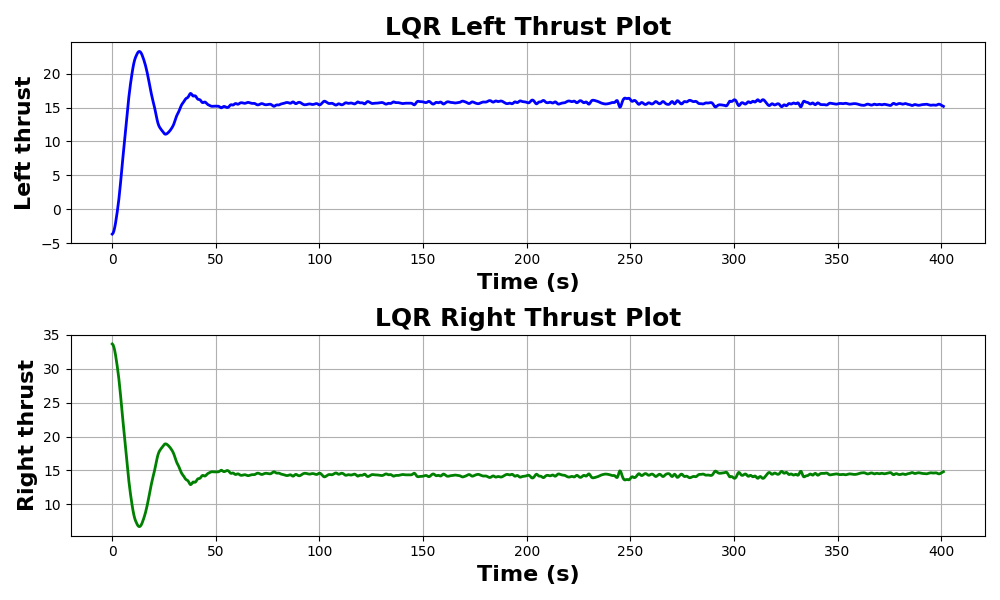} 
    \caption{Thrust command output for the LQR controller, showing the left  and right  thrust behavior over time in calm and clear sea conditions. }
    \label{fig:lqr_thrust}
\end{figure}

\begin{figure}[h]
    \centering
    \includegraphics[width=\columnwidth, page=1]{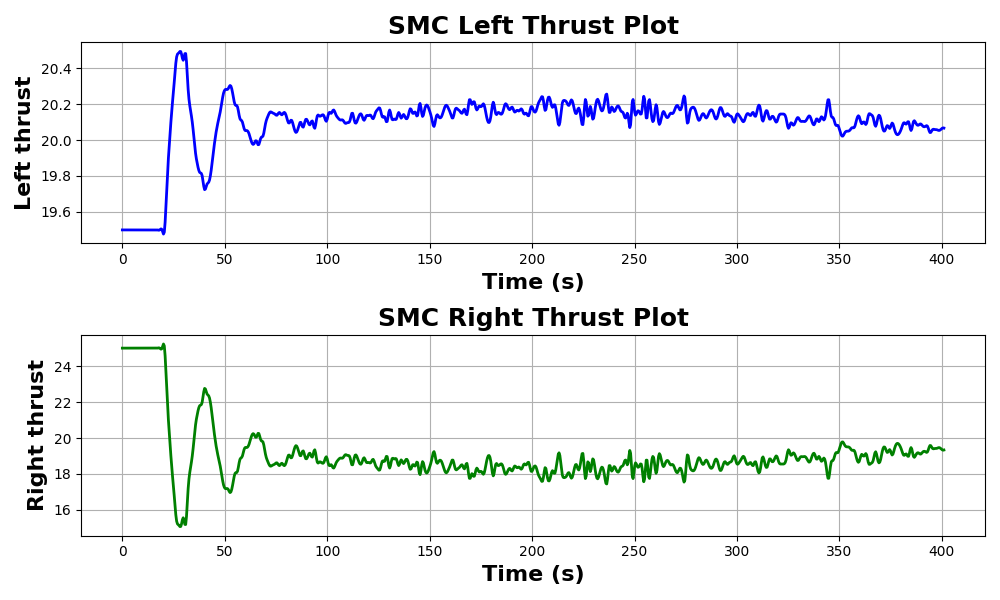} 
    \caption{Thrust command output for the SMC controller, showing the left  and right thrust behavior over time in calm and clear sea conditions.}
    \label{fig:smc_thrust}
\end{figure}

\begin{figure}[h]
    \centering
    \includegraphics[width=\columnwidth, page=1]{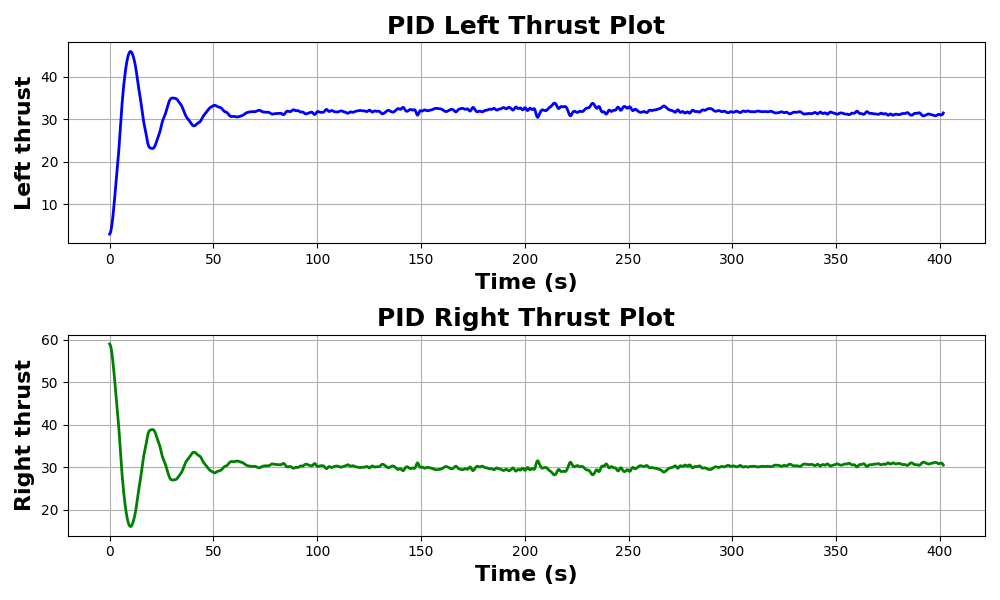} 
    \caption{Thrust command output for the PID controller, showing the left  and right thrust behavior over time in calm and clear sea conditions.}
    \label{fig:pid_thrust}
\end{figure}

\begin{figure*}[t]
\centering
\includegraphics[width=\linewidth, page=1]{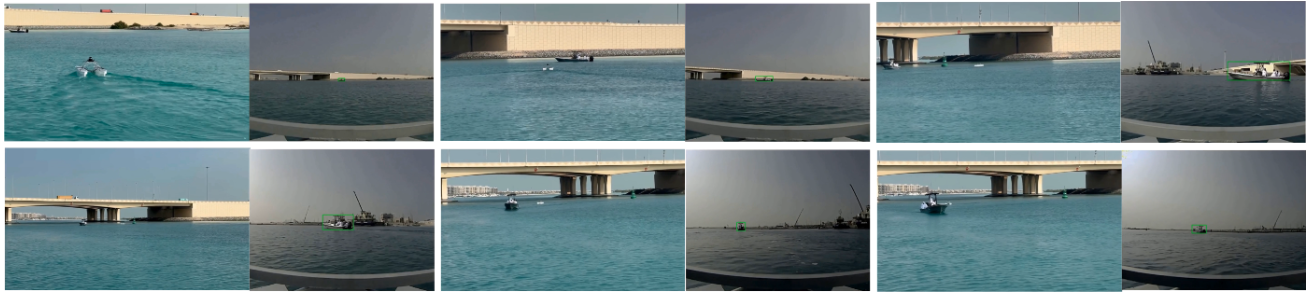}
    \caption{Snapshots of real-world tracking experiment in Saadiyat, Abu Dhabi, it captures the USV in various stages of target tracking. The figure demonstrates the camera output as the USV approaches and tracks the target in dynamic maritime conditions, with multiple angles and distances showing the system's robustness in real-time object detection and tracking.}
    \label{fig:realexp}
\end{figure*}
\subsection{Benchmarking on Real-World Dataset}
Similarly, for the real world data set, as shown in Table3, ToMP ranked as the top performer, achieving the highest AUC of 76.07 and the best OP50 and OP75 scores of 85.97 and 68.96, respectively. This indicates exceptional tracking accuracy and robustness at higher overlap thresholds. TaMOs and SeqTrack also demonstrated strong performance, with TaMOs achieving an AUC of 75.30 and OP50 and OP75 scores of 86.22 and 68.93, respectively, and SeqTrack achieving an AUC of 74.89 with OP50 and OP75 scores of 85.88 and 67.62, respectively. As shown in plots in Fig.~\ref{fig:realperformance}, these results highlight their superior tracking capabilities and consistency.
In contrast, SiamFC had shown the lowest performance with an AUC of 55.38 and lower OP50 and OP75 scores, indicating a less effective tracking accuracy. ATOM and DiMP showed improved results, with DiMP achieving the highest precision of 55.68 and Normalized Precision of 80.84, reflecting its strong tracking performance. In general, ToMP, TaMOs, and SeqTrack outperformed other models, providing robust tracking performance across the evaluated metrics.

The primary sensor in the proposed vision-based tracking framework is the camera; it is essential to the perception module for real-time target detection and monitoring in maritime environments. Although additional sensors, such as LiDAR, IMU, and DVL, are used to avoid obstacles and estimate state, the camera system is the main sensor for target tracking, particularly through vision-based trackers. However, in marine settings, cameras face significant challenges due to environmental factors, including bluer images due to water splashes on the camera and harsh weather, which can degrade image quality, reduce visibility, and impact tracking reliability. Tracker such as SeqTrack can handle such challenges; we have shown this by showing the tracker's performance under different environmental conditions, such as in calm sea conditions and in dust storms.

\section{Results and Discussions}
\subsection{Experimental Setup}
We use the MBZIRC maritime simulator, an open-source tool developed in C++ and Python. The simulator is designed to offer realistic maritime simulations by accurately implementing hydrodynamics and hydrostatics, which simulate the physical behavior of water and floating vessels.  Communication between the various modules in the simulator is handled through ROS2 Galactic, a middleware framework that enables seamless integration and real-time data exchange among different components of the system, such as sensors, control systems, and the user interface.
The simulator also has the ability to model environmental conditions. We can define various sea states, which include both wind direction and intensity, as well as the generation of waves to simulate real oceanic conditions. Additionally, it allows to model dust storms, which can further test the robustness of control and tracking algorithms under limited visibility and adverse weather conditions. As shown in Fig.~\ref{fig:scene}-A, the simulator is capable of rendering realistic visualizations of these environmental phenomena, such as dust storms that obscure visibility and challenge sensor reliability.

The real-world tests are a crucial step in bridging the gap between simulation and deployment, demonstrating the practical applicability of the developed system in real maritime environments.The real-world experiments were carried out in the coastal waters of Saadiyat Island, Abu Dhabi, to test and validate the performance of our tracking system in dynamic maritime conditions. For these trials, a motorboat was chosen as the moving target to simulate a realistic scenario for the USV's tracking capabilities. Fig.~\ref{fig:scene}-B
some examples of the tracking scenarios that we use in real demonstrations.

\subsection{Control Comparision}

Controllers required careful tuning to balance responsiveness and stability, particularly in dynamic conditions such as maritime environments. We tuned PID, SMC, and LQR for tracking, Fig.~\ref{fig:tuning} shows the controller tuning plots. The PID controller (red) shows a fast initial response, but experiences noticeable overshoot and subsequent oscillations before settling around zero error. The SMC (green) takes longer to reduce the initial error, but exhibits a smoother transient response with less oscillation, indicating robustness against dynamic changes. The LQR (blue) provides a more balanced performance, with a moderate initial error reduction and fewer oscillations during stabilization. .

We choose three trackers (SeqTrack, ToMP, and TaMOs) that show the best performance in benchmarking on the maritime dataset detailed in sec.~\ref{sec:benchmarking}. We use these trackers one by one in our tracking framework explained in sec.~\ref{sol:overview}. The results show that in real-time tracking applications, in calm and clear sea environments, all the above trackers shows similar performance. However, in dust storm, the SeqTrack outperforms ToMP, and TaMOs. SeqTrack shows more stable performance.  Fig.~\ref{fig:clearenvironment} and Fig~\ref{fig:dust}, show images captured by the USV onboard camera, illustrating how the tracker performs in a calm sea environment and under a dust storm.

We used SeqTrack to track the target in different scenarios for the controller evaluation. The scenarios that we used are; the target moving on a streight line and target moving on a triangular path. In calm sea conditions, all controllers show comparable performance as indicated in Figs.~\ref{fig:linecontrol} and~\ref{fig:tricontrol}. However, in a dust storm with the sea parameters  of wave gain = 0.5, wave period = 5.0, and wind linear velocity = \([1.5, -5.0, 0.0]\)
in both scenarios (line and triangle), the LQR outperforms the other two controllers. The LQR shows more stable and accurate tracking as shown in Fig~\ref{fig:dustline} and Fig~\ref{fig:triwind}.

The performance of the controller can also be evaluted by evaluating the thrust commands generated by the controllers while tracking. Fiq~\ref{fig:lqr_thrust}, ~\ref{fig:smc_thrust}, and~\ref{fig:pid_thrust} illustrate the thrust command generation for three control algorithms LQR, SMC, and PID,  during  calm and clear sea conditions. Each controller's performance is depicted by separate plots showing the left and right thrust outputs over time. In the case of the LQR controller, the thrust commands are smooth and consistent, showing minimal oscillations after the initial transient response, indicating a stable control output. The PID controller demonstrates more variability in the thrust output, with oscillations taking longer to settle. The SMC controller shows the noisiest thrust output, with significant fluctuations throughout the time period, indicating less smooth control compared to LQR and PID. Overall, the LQR controller generates the most stable and smooth thrust commands, while the SMC controller shows the weakest performance.

During the experiments, we observe significant differences in the controller’s performance, particularly when environmental conditions and sensor performance are considered. The performance of each controller is influenced by various factors, such as its inherent control characteristics and the reliability and limitations of the onboard sensors (such as DVL and IMU). In calm sea conditions, where sensor readings are stable and less noisy, the differences between controllers are relatively small. However, in scenarios with challenging conditions such as dust storms, high waves, and variable winds, sensor accuracy can be compromised. This required the control algorithms to be robust enough to compensate for sensor limitations. The LQR controller consistently demonstrates better tracking stability and control precision in varying environmental conditions. 
The LQR controller outperforms both SMC and PID by providing smoother thrust commands and a quicker return to stability, even when sensor inputs are noisy or degraded. This stability minimizes the noise of the control system, which is especially beneficial when the sensor data is unreliable, ultimately enhancing the USV’s capacity to maintain a steady lock on the moving target. The synergy between SeqTrack and the LQR controller presents a practically robust solution for maritime tracking applications. The combination of precise vision-based tracking (SeqTrack) and the adaptive stability of the LQR controller shows robust target tracking in real-world maritime scenarios, where environmental disturbances can undermine sensor accuracy. This integrated approach effectively addresses the dual challenges of maintaining accuracy and stability, underscoring the practical relevance of this set-up for real-world deployment in maritime search-and-rescue operation, environmental monitoring, and autonomous navigation applications.

During real-world experiments, we chose SeqTrack for target tracking, as it consistently demonstrated the best performance in both simulation environments and on real-world datasets. Among the various controllers we tested, we used the LQR  controller for tracking because of its superior stability and control precision.
Fig.~\ref{fig:realexp} illustrates a sequence of snapshots capturing the USV's tracking behavior alongside the corresponding tracker's output. These images show the position and trajectory during the tracking process, showing how effectively the system maintains the target lock under actual sea conditions. The combination of SeqTrack and LQR was highly effective, providing more accurate, stable, and smooth tracking compared to other control strategies tested. The LQR controller, in particular, minimized oscillations and maintained steady control, while SeqTrack ensured precise target localization and tracking.

To deploy the developed tracking solution in real-time maritime environments, there is a trade-off between accuracy and computational efficiency for tracking and control. It can affect the system's responsiveness and overall performance. High-accuracy algorithms, such as the LQR controller and SeqTrack tracker used in our experiments, offer precise tracking and stability under varying sea conditions, as demonstrated in calm and adverse environments. However, achieving this level of accuracy requires intensive computation, which can drain system resources, particularly in real-time applications where computational power is limited. 
For example, the SeqTrack tracker showed robust performance in dust storm conditions, maintaining higher accuracy in adverse visibility compared to ToMP, TaMOs, and other trackers. However, SeqTrack is computationally intensive due to the transformer-based architecture. In real-time systems, where on-board processing capabilities may be constrained, this could introduce delays in control feedback loops, potentially affecting the correctness of control adjustments. The LQR controller also demonstrates the trade-off between accuracy and processing speed, providing smooth and stable control with minimal oscillations, as shown in Fig.~\ref{fig:lqr_thrust}. Although its computational demands are moderate compared to the SMC and PID controllers, achieving this stability can be computationally intensive if it is continuously optimized in real time. 
All the trackers and controllers discussed show comparable performance in calm sea and in clear visibility conditions. Therefore, any of the controllers and trackers can be used. However, in harsh sea conditions with reduced visibility, SeqTrack and LQR perform the best but require a higher computational cost. 

\section{Conclusion}
This research introduced a vision-guided real-time object tracking framework for USVs. integrating state-of-the-art vision-based trackers with low-level control systems to achieve precise tracking in dynamic maritime
environments.  We benchmarked the performance of six advanced deep learning-based trackers and assessed the robustness of various control algorithms. We validated the framework through simulations and real-world sea experiments to highlight its effectiveness in handling dynamic maritime conditions.
The results show that SeqTrack, a Transformer-based tracker, performed best in adverse conditions, such as
dust storms, while the LQR controller provided the most stable and robust tracking performance.

\section*{Decleration}
During the preparation of this work the author(s) used ChatGPT and Bard in order to improve language and readability. After using this tool/service, the author(s) reviewed and edited the content as needed and take(s) full responsibility for the content of the publication.
\bibliographystyle{IEEEtran}
\bibliography{References}

\balance

\end{document}